\documentclass[times, review, 10pt]{elsarticle}

\usepackage[letterpaper,top=4.3cm,right=4.8cm,bottom=4.3cm,left=4.8cm]{geometry}
\usepackage{amsmath,amssymb}
\usepackage{booktabs}
\usepackage{float}
\usepackage[ruled,linesnumbered,vlined]{algorithm2e}
\biboptions{numbers,sort&compress}
\usepackage{caption}
\usepackage{etoolbox}
\usepackage[hidelinks]{hyperref}
\hypersetup{
  pdftitle={FedHUG: Federated Heterogeneous Unsupervised Generalization for Remote Physiological Measurements},
  pdfauthor={Xiao Yang, Dengbo He, Duoxun Tang, Jiyao Wang, Kaishun WU}
}
\DeclareUrlCommand\path{\urlstyle{same}}

\newcommand{\editorialsmall}{\fontsize{8}{10}\selectfont}
\makeatletter
\renewcommand\footnotesize{%
  \@setfontsize\footnotesize{8}{10}%
  \abovedisplayskip 6pt plus 2pt minus 4pt
  \abovedisplayshortskip 0pt plus 1pt
  \belowdisplayshortskip 3pt plus 1pt minus 2pt
  \def\@listi{\leftmargin\leftmargini
    \topsep 3pt plus 1pt minus 1pt
    \parsep 2pt plus 1pt minus 1pt
    \itemsep \parsep}%
  \belowdisplayskip \abovedisplayskip}
\patchcmd{\pprintMaketitle}{\Large\@title}
  {\fontsize{14}{17}\selectfont\@title}{}
  {\PackageError{editor-format}{Title font patch failed}{Check elsarticle.cls.}}
\patchcmd{\pprintMaketitle}{\normalsize\elsauthors}
  {\editorialsmall\elsauthors}{}
  {\PackageError{editor-format}{Author font patch failed}{Check elsarticle.cls.}}
\patchcmd{\pprintMaketitle}{\footnotesize\itshape\elsaddress}
  {\editorialsmall\itshape\elsaddress}{}
  {\PackageError{editor-format}{Affiliation font patch failed}{Check elsarticle.cls.}}
\patchcmd{\emailauthor}{\ttfamily}{\rmfamily}{}
  {\PackageError{editor-format}{Email font patch failed}{Check elsarticle.cls.}}
\makeatother
\DeclareCaptionFont{editorialeight}{\editorialsmall}
\SetAlCapFnt{\editorialsmall}
\SetAlCapNameFnt{\editorialsmall}

\DeclareUnicodeCharacter{2191}{\ensuremath{\uparrow}}
\DeclareUnicodeCharacter{2193}{\ensuremath{\downarrow}}

\usepackage{setspace}
\begin{document}
\begin{frontmatter}

\title{FedHUG: Federated Heterogeneous Unsupervised Generalization for Remote Physiological Measurements}

\author[aff1]{Xiao Yang\fnref{equal}}
\ead{xyang856@connect.hkust-gz.edu.cn}

\author[aff2]{Dengbo He\fnref{equal}}
\ead{dengbohe@hkust-gz.edu.cn}

\author[aff3]{Duoxun Tang}
\ead{tdx25@mails.tsinghua.edu.cn}

\author[aff2]{Jiyao Wang\corref{cor1}}
\ead{jwanggo@connect.ust.hk}

\author[aff1]{Kaishun WU}
\ead{wuks@hkust-gz.edu.cn}

\cortext[cor1]{Corresponding author.}
\fntext[equal]{These authors contributed equally to this work.}

\affiliation[aff1]{organization={Info Hub, The Hong
Kong University of Science and Technology (Guangzhou)},
  city={Guangzhou}, postcode={511453}, country={China}}
\affiliation[aff2]{organization={Systems Hub, The Hong
Kong University of Science and Technology (Guangzhou)},
  city={Guangzhou}, postcode={511453}, country={China}}
\affiliation[aff3]{organization={Shenzhen International Graduate School, Tsinghua University}, city={Shenzhen}, postcode={518055}, country={China}}

\begin{abstract}
\doublespacing
  Remote physiological measurement gained wide attention, while it requires collecting users' privacy-sensitive information, and existing contactless measurements still rely on labeled client data. This presents challenges when we want to further update real-world deployed models with numerous user data lacking labels. To resolve these challenges, we instantiate a new protocol called Federated Unsupervised Domain Generalization (FUDG) in this work. Subsequently, the \textbf{Fed}erated \textbf{H}eterogeneous \textbf{U}nsupervised \textbf{G}eneralization (\textbf{FedHUG}) framework is proposed and consists of: (1) Minimal Bias Aggregation module dynamically adjusts aggregation weights based on prior-driven bias evaluation to cope with heterogeneous non-IID features from multiple domains. (2) The Global Distribution-aware Learning Controller parameterizes the label distribution and dynamically manipulates client-specific training strategies, thereby mitigating the server-client label distribution skew and long-tail issue. The proposal shows superior performance across state-of-the-art techniques in estimation with either RGB video or mmWave radar. The code will be released.
\end{abstract}

\begin{keyword}
rPPG \sep Federated Generalization \sep Remote Physiological Measurements
\end{keyword}

\end{frontmatter}

\doublespacing

\section{Introduction}
\label{sec:intro}
Traditional physiological measurements, like electrocardiograms (ECG) and blood volume pulse (BVP), are accurate but face challenges such as high costs and inconvenient wearability \cite{wang2025physmle}. Therefore, remote physiological measurement (RPM) using non-contact multimedia devices, such as RGB cameras \cite{verkruysse2008remote} and millimeter-wave (mmWave) sensors \cite{hu2024contactless}, offers a promising alternative. This technology has shown potential in human-computer interaction \cite{wang2024multi}, affective computing \cite{lu2024gpt}, and state monitoring \cite{wang2024efficient}. Early studies \cite{niu2019rhythmnet} have shown that deep learning methods outperform traditional approaches in complex environments. However, DL models trained on a limited amount of data still struggle to generalize effectively across real-world scenarios \cite{wang2023hierarchical} due to differences in factors (e.g., illumination, skin color).

Recently, researchers paid increasing attention to mitigating domain shifts and enhancing the model's generalization ability \cite{wang2024tim, yang2025consistencyenhancetesttimeadaptation}. However, there are obstacles to deploying generalizable methods, primarily due to the privacy-sensitive nature of physiological data. The collection of users' private information in deployment environments significantly reduces user acceptance \cite{wang2024evaluating}. Liu et al. \cite{liu2022federated} proposed using federated learning (FL) for RPM to keep users' data local during model training. However, existing methods still require labeled data, which is unavailable on the user side. Additionally, domain shifts occur across different scenarios and users \cite{tang2023mmpd}. Updating the model effectively while ensuring privacy and mitigating performance degradation due to domain shifts is challenging in deployed environments.

\begin{figure}[ht] 
    \centering
    \includegraphics[width=0.9
    \textwidth]{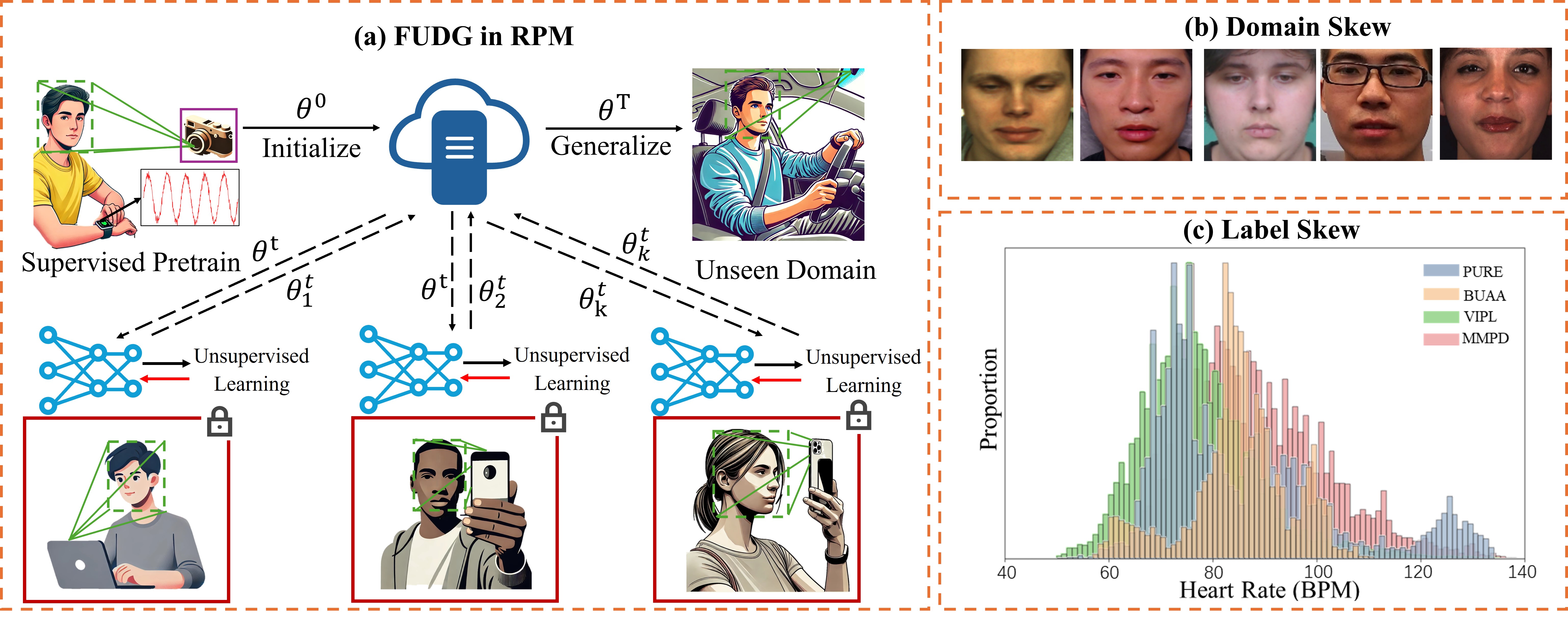} 
    \caption{ (a) Overview of FUDG. We take the video-based physiological estimation as an example. (b) Domain Skew. Significant feature shifts under varying user attributes and environmental factors.(c) Label Skew. Inherent long-tail and imbalanced distribution of phys-labels across datasets (i.e., PURE \cite{song2021pulsegan}, BUAA \cite{xi2020image}, VIPL \cite{lu2021dual}, MMPD \cite{tang2023mmpd}).}
    \label{fig:FedHUG}
\end{figure}

Therefore, we begin by instantiating the Federated Unsupervised Domain Generalization (FUDG) protocol \cite{pourpanah2024federated} in RPM. Our goal is to fine-tune a pre-trained model in an unsupervised manner during the FL process to enhance generalization. As shown in Fig. \ref{fig:FedHUG}(a), a pre-trained network is deployed on the global server and shared with clients collecting unlabeled inputs from diverse environments. Through unsupervised learning, client model parameters are uploaded to the server for aggregation, enabling the global model to adapt to changing scenarios.

To achieve this goal, we need to tackle two challenges in FUDG: (1) Biases from data collection devices and user differences \cite{aslami2024comparative} make domain skew in feature distribution, as shown in Fig. \ref{fig:FedHUG}(b), complicating generalization during FL (2) The long-tail phenomenon in global physiological labels (phys-labels) distribution \cite{wang2024rppg} exists and shows distributional heterogeneity across clients \cite{ma2025geometric}. Particularly, due to limitations in privacy concerns (neither ground-truth nor pseudo-label should be transmitted), and the unavailability of ground-truth, we cannot directly acquire the global distribution. Local learning with label skew will result in inconsistency in the update direction during model aggregation, which further affects the convergence and performance of the global model \cite{zhang2022federated}. Meanwhile, due to the strict privacy constraints in FUDG, clients blindly overfit to their local majorities without a global consensus. For example, as explicitly revealed in Fig. \ref{fig:FedHUG}(c), the tail region of one domain (e.g., high HRs in PURE) can act as the dominant head region in another (e.g., MMPD). Such drastic distribution mismatches lead to severely conflicting gradients, effectively contributing no learning benefit toward the global minority states.

These challenges can all be seen as specific manifestations of statistical heterogeneity \cite{ye2023heterogeneous}. To address it, we propose a novel \textbf{Fed}erated \textbf{H}eterogeneous \textbf{U}nsupervised \textbf{G}eneralization (\textbf{FedHUG}) framework to enhance model generalization during the unsupervised federated learning. Specifically, we introduce a novel Minimal Bias Aggregation (\textbf{MBA}) method for the FUDG task. With the prior-based bias assessment, it assigns higher weights to clients with smaller biases. It evaluates the distance between the learned semantics from different clients and the ideal semantic-invariant point, thus guiding the updating direction of the global model toward the optimization point with less bias. Additionally, to resolve the semantic distributional shift between domains and agnostic global distribution, we designed a Global Distribution aware Learning Controller (\textbf{GDLC}), which precisely regulates the training process on the client side through implicit semantic distribution modeling on the server, without any privacy-concern information transmission. Based on it, FedHUG can encourage clients to enhance learning of tail samples, and mitigate the negative impact of long-tail semantic distributions. Our contributions are outlined as follows:

\begin{itemize}
    \item As far as we know, we are the first to introduce the concept of federated unsupervised domain generalization and establish a new FUDG protocol for remote physiological measurement.
    \item A novel FedHUG framework is proposed to mitigate heterogeneity in feature and semantic space across multi-domains in a federated, unsupervised way.
    \item  We establish a large-scale benchmark for the RPM task under the multi-source FUDG protocol with both video and mmWave-based measurements.
    \item Extensive experiments demonstrate significant improvements and effectiveness in our method compared to SOTA approaches.
\end{itemize}

\section{Related Work}
\subsection{Generalizable Remote Physiological Measurement} Early studies \cite{de2013robust, wang2016algorithmic} estimated BVP signals and HR by capturing changes in the color space signals of facial videos, a technique known as remote photoplethysmography (rPPG). At the same time, wireless signals, such as ultra-wideband (UWB) and mmWave signals, have also been used in contactless cardiac health monitoring. However, these methods are highly sensitive to environmental changes, variations in light intensity, and slight movements, making them difficult to use in complex scenarios. With advancements in deep learning, an increasing number of DL methods have been proposed and achieved better performance by improving model structures \cite{yu2019remote} and feature constraints \cite{sun2024contrast} tailored to physiological signals. Despite these methods showing good performance on a single dataset, they struggle with generalization in real-world environments \cite{wang2023hierarchical}. To address these issues, recent research has started focusing on domain generalization strategies \cite{wang2025towards, wang2025physmle} aimed at enhancing the model's generalization capability. However, more unlabeled data is collected from various environments and users in the real world. Data is often spread across local devices, and privacy concerns limit the central server's access, hindering the model's scalability and generalization.

\subsection{Federated Learning}

FL \cite{mcmahan2017communication} facilitates decentralized training without raw data exposure. Various FL paradigms have since emerged to mitigate data heterogeneity and communication bottlenecks  \cite{li2020federated}. Given the stringent privacy requirements and vulnerability to attacks in healthcare, FL has become a cornerstone for multi-source coordination. Driven by escalating privacy concerns and potential attack vulnerabilities in healthcare \cite{tang2025featurefool}, FL has gained massive traction for multi-source coordination \cite{antunes2022federated}.  

In the RPM domain, Liu et al. \cite{liu2022federated} pioneered an FL-based rPPG method, but it heavily relies on annotated client data. Furthermore, while Federated Unsupervised learning (FU) \cite{zhang2023federated} eliminates the need for labels, existing FU frameworks largely assume identical data distributions and lack generalization capabilities across unseen domains. More recently, a novel protocol called FUDG \cite{pourpanah2024federated} has been introduced to extract general representations from decentralized, unlabeled datasets. Unlike traditional FU, FUDG targets both non-IID data and the domain shift problem in an unsupervised manner.

\section{Methodology}
\subsection{Problem Formulation}
\label{subsection: Problem Formulation}

To improve RPM generalization in unseen domains using unlabeled data, we adopt the FUDG protocol with $K$ decentralized clients $\{C_i\}_{i=1}^K$. For privacy, raw sensory data are kept strictly local and immediately transformed into spatio-temporal representations (e.g., STMaps and Range Matrices) \cite{niu2019rhythmnet, wu2025cardiacmamba}. Though visually de-identified, they retain sensitive cardiovascular biomarkers, making localized processing imperative for healthcare compliance. Thus, each client's private dataset is denoted as $\mathcal{D}_i = \{x_i^n\}_{n=1}^{N_i}$ with $N_i$ samples.

Assume that data distributions across clients are statistically heterogeneous, denoted as $P_i(X) \neq P_j(X)$ for $i \neq j$. During federated learning, the model parameterized by $\theta$ acts as a mapping function to extract abstract physiological features $v \in \mathbb{R}^d$ from inputs, and subsequently predicts the physiological states. The ultimate goal is to learn a globally generalized representation capable of minimizing the expected loss on an unseen target domain $\mathcal{D}_u$ with distribution $P_u$, where $P_u \neq P_i$:

\begin{equation}\min_{\theta} \mathbb{E}{x \sim P_u} [\mathcal{L}(\theta; x)]
\label{eq:global_goal}
\end{equation}

where $\mathcal{L}(\cdot)$ denotes the unsupervised loss function. Since the FUDG protocol strictly prohibits access to the unseen domain data $\mathcal{D}_u$ during training, each client can only optimize its local parameters $\theta_i$ via Empirical Risk Minimization on its own empirical distribution:

\begin{equation}
\min_{\theta_i}\,\mathbb{E}_{x \sim P_i}\!\left[\mathcal{L}(\theta_i; x)\right]
\;\approx\;
\min_{\theta_i}\,\frac{1}{N_i}\sum_{n=1}^{N_i}\mathcal{L}(\theta_i; x_i^n).
\label{eq:local_goal}
\end{equation}

At communication round $t$, after all clients update their local parameters to $\theta_i^t$, the central server aggregates them into a new global model $\theta^{t+1}$. The standard FedAvg \cite{mcmahan2017communication} assigns aggregation weight $p_i$ simply proportional to the local dataset size:\begin{equation}\theta^{t+1} = \sum_{i=1}^K p_i \theta_i^t, \quad \text{where} \quad p_i = \frac{N_i}{\sum_{j=1}^K N_j}\label{eq:fedavg}\end{equation}

However, this naive aggregation scheme is heavily vulnerable in the FUDG setting. It blindly treats all local updates as equally reliable (scaled only by quantity $N_i$), completely ignoring the varying degrees of domain-specific bias (Feature Skew) and the severe heterogeneity of physiological long-tail distributions (Label Skew) across clients. To conquer these bottlenecks, we propose the FedHUG framework, which incorporates a Minimal Bias Aggregation mechanism to dynamically evaluate local semantic robustness (Section \ref{subsection: Minimal Bias Aggregation}), and a Global Distribution-Aware Learning Controller to regulate local representation space implicitly (Section \ref{subsection: Global Distribution Aware Learning Controller}).

\subsection{Multi-Modal Representation and Physiological Augmentation}
\label{subsection: Multi-Modal Representation and Physiological Augmentation}

\subsubsection{Multi-Modal Representation Construction}

For RGB videos, we transform each facial sequence into a Spatial-Temporal Map (STMap) to reduce computational cost and emphasize changes in facial skin brightness rather than environmental variation, following \cite{lu2023neuron}. Specifically, FAN \cite{bulat2017far} is used to detect 2D facial landmarks. The average RGB values within each facial region of interest (ROI) are then arranged over time to form $x_i^n \in \mathbb{R}^{T \times W \times C}$, where $T$ is the number of frames, $W$ is the number of facial ROIs, and $C=3$ corresponds to the RGB channels. Figure \ref{fig:representation_augmentation}(a) illustrates this construction.

For mmWave measurements, the raw radar signal is processed by range-angle-Doppler FFTs following \cite{wu2025cardiacmamba}. The resulting range matrix is represented by the same tensor notation $x_i^n \in \mathbb{R}^{T \times W \times C}$, where $T$ denotes time frames, $W$ indexes the retained spatial bins within the region of interest, and $C=2$ stores the real and imaginary components of the complex radar signal. This common tensor organization allows the same federated learning protocol and consistency analysis to operate on both sensing modalities.

\subsubsection{Physiologically Consistent Spatio-Temporal Augmentation}

\begin{figure}[htbp]
    \centering
    \begin{minipage}[c]{0.67\textwidth}
        \centering
        \includegraphics[width=\linewidth]{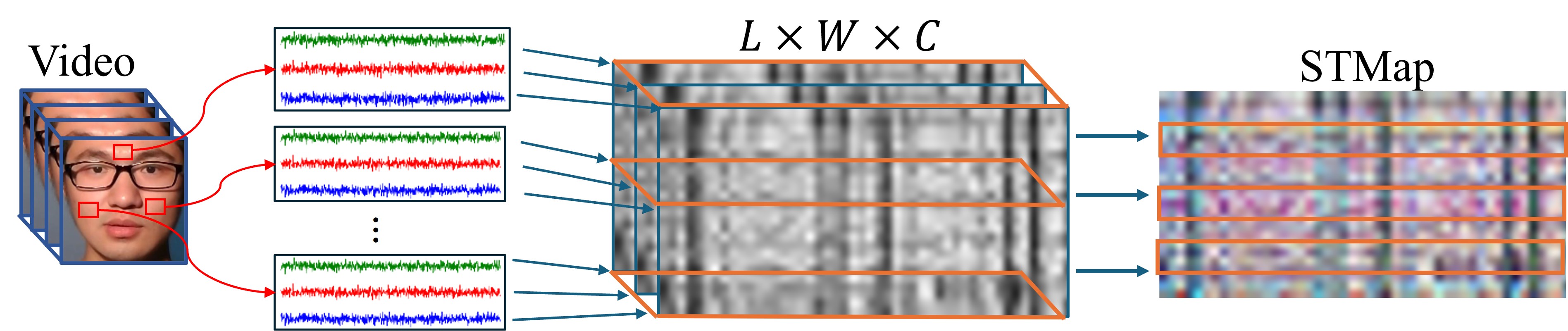}
        \par\textbf{(a)}
    \end{minipage}\hfill
    \begin{minipage}[c]{0.30\textwidth}
        \centering
        \includegraphics[width=\linewidth]{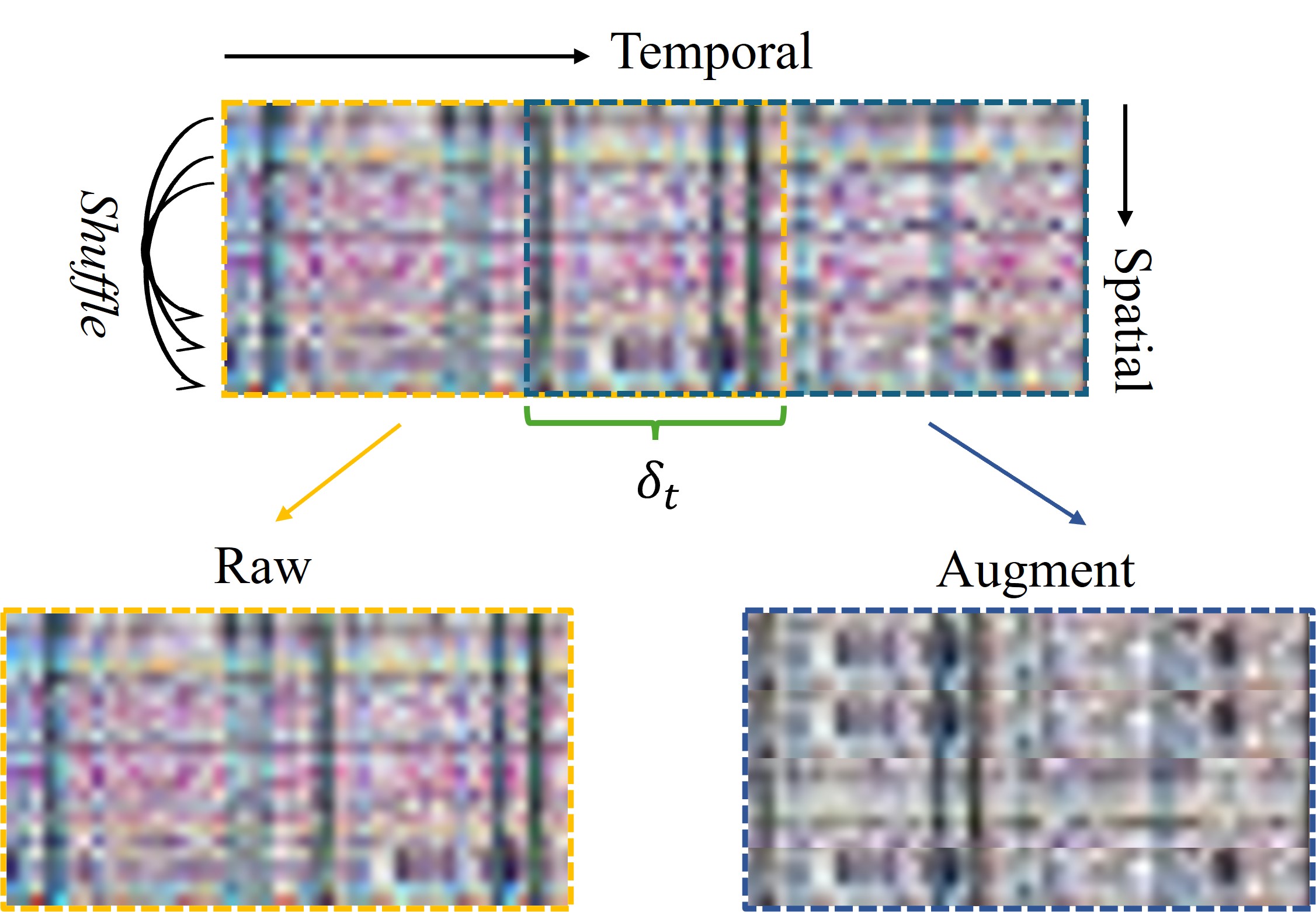}
        \par\textbf{(b)}
    \end{minipage}
    \caption{RGB representation construction and physiologically consistent augmentation. (a) Production of an STMap from a facial RGB video using region-wise color traces. (b) Unified spatio-temporal augmentation, illustrated with an STMap: temporal sliding introduces a phase shift $\delta_t$, while spatial permutation disrupts structural and environmental noise. The same temporal and spatial operations are applied to mmWave range matrices.}
    \label{fig:representation_augmentation}
\end{figure}

MBA requires two views that differ in domain-specific appearance or structure while preserving the underlying physiological rhythm. We therefore apply a unified spatio-temporal augmentation strategy to both RGB-based STMaps and mmWave-based range matrices. As summarized in Figure \ref{fig:representation_augmentation}(b), an original input $x_i^n$ is transformed into an augmented view $a_i^n$ through temporal shifting and spatial shuffling.

For temporal augmentation, we introduce a random offset $\delta_t \sim \mathcal{U}(0,\Delta)$ following a discrete uniform distribution, with $\Delta=30$ frames. For $x_i^n \in \mathbb{R}^{T \times W \times C}$, the shift is applied along the time dimension as
\begin{equation}
    x_{\tau,w,c}^{a} \leftarrow x_{\tau+\delta_t,w,c}^{n}.
    \label{eq:temporal_augmentation}
\end{equation}
This operation simulates time delays, user micro-movements, and sampling inconsistencies across heterogeneous devices. It changes the phase of the underlying BVP signal without altering its intrinsic frequency.

For spatial augmentation, the bands along the spatial dimension $W$ are randomly permuted. For RGB STMaps, $W$ represents facial ROIs; shuffling disrupts identity-specific facial topology and illumination patterns, encouraging the network to model global chrominance variation rather than a particular skin region. For mmWave range matrices, $W$ represents spatial range bins; permutation simulates variations in subject distance, posture, and environmental multipath reflections, thereby reducing sensitivity to local spatial clutter.

The joint transformation introduces a phase delay and structural variation between $x_i^n$ and $a_i^n$. A robust model should nevertheless extract consistent cardiac semantics from both views. This physiological invariance provides the basis for the SSM consistency score used by MBA in Section \ref{subsection: Minimal Bias Aggregation}.

\subsection{Minimal Bias Aggregation}
\label{subsection: Minimal Bias Aggregation}

\begin{figure}[ht] 
    \centering
    \includegraphics[width=0.90\textwidth]{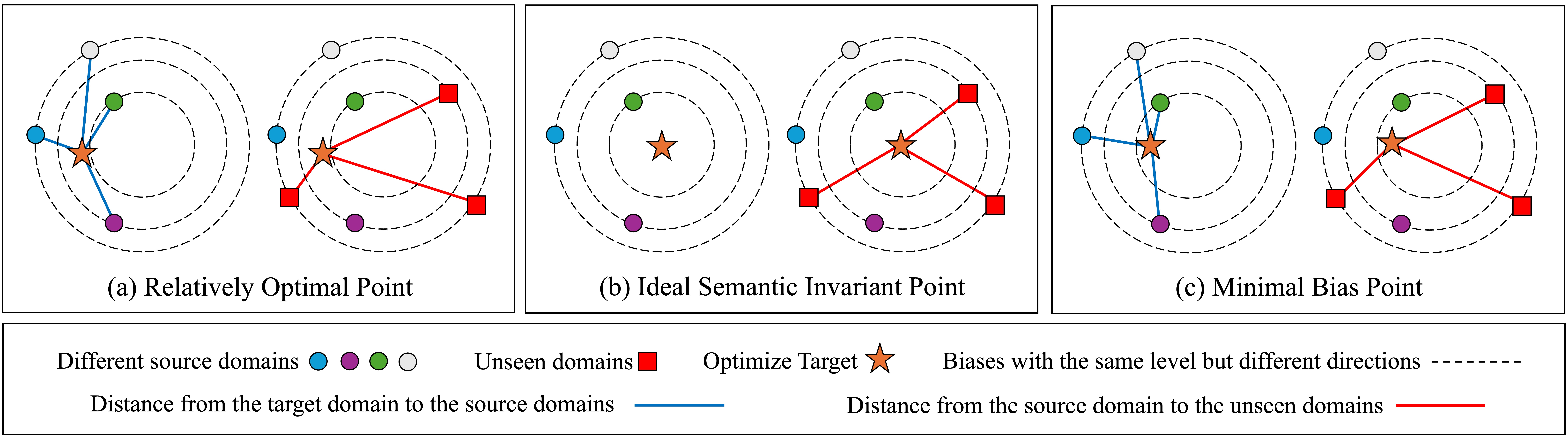} 
    \caption{Illustration of three different aim domains. Here, the dashed circle’s center represents the semantic features, and the closer a domain is to the center, the less interference it contains.}
    \label{fig:Ideal Domain}
\end{figure}

According to \cite{zhang2025advancing}, the fitted model can be divided into two parts: (1) inherent semantic information, indicating a strong correlation between facial color changes and physiological states; (2) bias terms caused by various types of noise, which is considered one of the key factors contributing to statistical heterogeneity \cite{li2020federated, luo2021no}. Previous DG methods \cite{ wang2023hierarchical} typically encourage the model to generalize to the point where the sum of distances to each source domain is minimized, as shown in Fig. \ref{fig:Ideal Domain}(a). However, it is merely a Relatively Optimal Point within a known distribution \cite{wang2024tim}, and is still troubled by bias in source domains. The ideal approach is to align the model with a fair distribution that is free from bias and contains only invariant semantic information \cite{li2018domain}. As shown in Fig. \ref{fig:Ideal Domain}(b), given such an Ideal Semantic Invariant Point, different domains influenced by bias are scattered around an ideal point, with those having smaller biases closer to it. However, due to varying degrees of uncontrollable factors in data collection and individual biases,  the ideal semantic point is agnostic, and we cannot quantify bias impact.

To circumvent the intractable nature of explicitly modeling the ideal point, we propose to implicitly constrain the model using robust prior knowledge. Specifically, the spatio-temporal consistency inherent in physiological states \cite{sun2022contrast,sun2024contrast} dictates that valid BVP signals should maintain morphological stability regardless of spatial or temporal perturbations. Any significant deviation under such perturbations is therefore likely attributable to external noise or identity-specific bias. Building upon this, the augmentation defined in Section \ref{subsection: Multi-Modal Representation and Physiological Augmentation} transforms the original input \(x_i^{n}\) into an augmented view \(a_i^{n}\) \cite{wang2025physmle}. The extracted physiological semantics (i.e., the underlying HR rhythm) from \(x_i^{n}\) and \(a_i^{n}\) should remain highly consistent. Conversely, a pronounced semantic discrepancy between them signals a violation of this consistency, indicating a higher degree of bias in the current representation.

The subsequent challenge lies in accurately measuring this discrepancy to quantify the impact of bias. Since spatio-temporal augmentations inevitably introduce temporal offsets, a phase delay often exists between the predicted BVP signals of \( x_i^{n}\) to \(a_i^{n}\). This renders standard point-to-point alignment metrics (e.g., Pearson correlation) inapplicable. To address this, we draw inspiration from the Self-Similarity Matrix (SSM) \cite{wang2026align, zhang2024self}. While existing RPM studies primarily utilize SSM as a robust feature representation to mitigate label phase conflicts and preserve temporal physiological rhythms, we elegantly repurpose it in our unsupervised FUDG setting as a translation-invariant surrogate metric. 

\begin{figure}[ht] 
    \centering
    \includegraphics[width=0.85\textwidth]{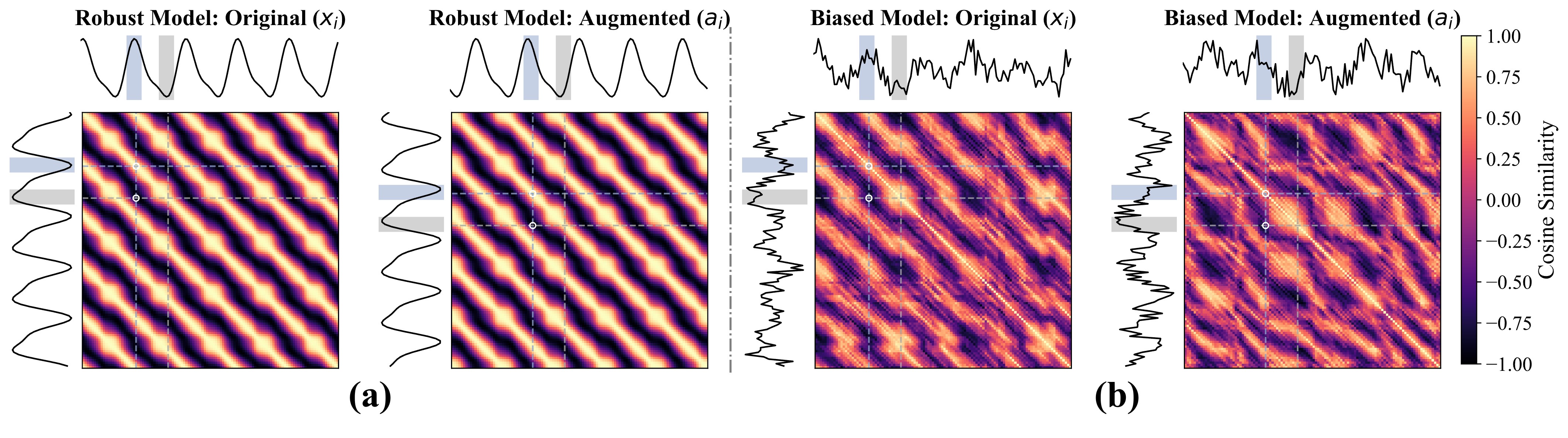} 
    \caption{Minimal bias evaluation via SSM generation. (a) The robust model yields highly consistent translation-invariant SSMs despite the phase shift between the original ($x_i$) and augmented ($a_i$) views. (b) The biased model succumbs to domain-specific noise, resulting in corrupted and inconsistent SSM textures. This visual discrepancy enables the score $s$ to accurately quantify model bias.}
    \label{fig:SSM}
\end{figure}

As visually detailed in Fig. \ref{fig:SSM}, the 1D signals are mapped to an auto-correlation matrix, where intersections of highly correlated periodic windows (e.g., peak-to-peak) naturally yield high similarities. This mechanism makes SSM strictly robust against absolute phase shifts while completely retaining the underlying physiological frequency. Specifically, for a robust model (Fig. \ref{fig:SSM}(a)), even if the two signals exhibit a phase delay, their cardiac rhythms yield highly consistent SSM patterns. Conversely, if a model is severely biased by domain noise (Fig. \ref{fig:SSM}(b)), its semantic representation collapses under perturbations, leading to corrupted and inconsistent SSM textures across views. Formally, for the $n$-th sample $x_i^n$ on client $i$, we employ a sliding window (length $l$, step 1) to extract segment \(U = \{{u_1, u_2, ...,u_{L-l+1}}\}\) from the BVP signals. By iteratively computing the standard cosine similarity between each segment pair $(u_p^n, u_q^n)$, we construct the translation-invariant SSM $\mathcal{M} \in \mathbb{R}^{(L-l+1)\times(L-l+1)}$ for the BVP signal.

Lastly, the client evaluates the semantic invariance over its entire local dataset $\mathcal{D}_i$ by computing the averaged consistency score $s_i$. As formulated in Eq. (\ref{eq:COSINE}), we utilize the Frobenius inner product to calculate the cosine similarity between the original SSM $\mathcal{M}^n$ and the augmented SSM $\mathcal{M}_a^n$:
\begin{equation}
\textbf{$s_i = \frac{1}{N_i} \sum_{n=1}^{N_i} \frac{\langle \mathcal{M}^n, \mathcal{M}_a^n \rangle_F}{\|\mathcal{M}^n\|_F \|\mathcal{M}_a^n\|_F}$}
\label{eq:COSINE}
\end{equation}

\begin{figure}[ht] 
    \centering
    \includegraphics[width=0.90\textwidth]{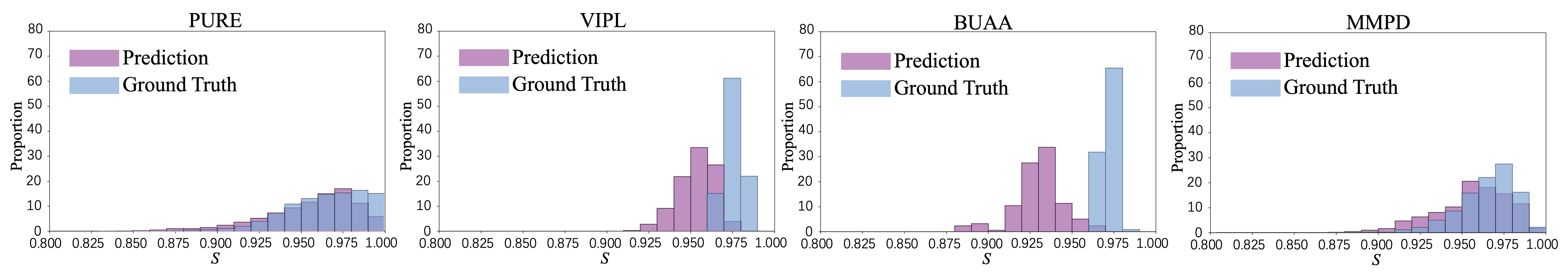} 
    \caption{Illustration of the distribution of \(s\) in different datasets.}
    \label{fig:GAP}
\end{figure}

Statistical analysis further validates our hypothesis. As shown in Fig. \ref{fig:GAP}, while the consistency score \(s\) calculated from the ground-truth BVP signals exhibits slight variance—attributable to inherent physiological fluctuations and inevitable measurement noise—it forms a tightly concentrated distribution closely approaching 1. This empirical distribution establishes a realistic benchmark for spatio-temporal consistency. In contrast, the similarity distribution produced by the baseline model is notably shifted towards lower values with a significantly wider variance. This substantial discrepancy demonstrates that pre-trained models struggle to maintain semantic invariance and are instead highly susceptible to domain-specific biases. Ultimately, by analyzing and optimizing the SSM consistency across domains, we establish an effective surrogate metric. This metric guides the model to mitigate bias and approximate the semantic invariant distribution, bypassing the need to explicitly locate the intractable ideal point.

Given these, we propose the Minimal Bias Aggregation (\textbf{MBA}) to encourage the model to focus more on data that is closer to the ideal feature distribution during training. Since the new optimization target point has minimal bias compared to all domains and the previous relatively optimal point, we consider it as the Minimal Bias Point (as shown in Fig. \ref{fig:Ideal Domain}(c)).
Specifically, not only are parameters \(\theta_i\) uploaded from each client, , but each client also transmits its dataset-level consistency score $s_i$ to the server. All the scores $S = \{s_1, s_2, \dots, s_K\}$ are converted into scaling factors via a Softmax function with a temperature coefficient $\tau$.  Consequently, the clients' parameters with smaller bias (higher $s_i$) will contribute more to the updates on the server, while the contribution of clients with higher bias will be diminished, as follows:

\begin{equation}
    \omega_i^t = \frac{p_i\exp(s_i/\tau)}{\sum_{j=1}^{K}p_j\exp(s_j/\tau)},
    \qquad
    \theta^{t+1} = \sum_{i=1}^{K}\omega_i^t\theta_i^t
    \label{eq:mba}
\end{equation}

where $p_i$ is the original data volume proportion defined in Eq. (\ref{eq:fedavg}), and $\omega_i^t$ is the normalized aggregation weight satisfying $\sum_{i=1}^{K}\omega_i^t=1$.

\subsection{Global Distribution Aware Learning Controller}
\label{subsection: Global Distribution Aware Learning Controller}

Previous studies \cite{wang2024rppg} have shown that the distribution of the phys-labels in the same dataset typically follows a long-tail pattern, with significant distribution differences between datasets, which is considered a manifestation of distribution heterogeneity \cite{tomitani2021regional}. Moreover, in the FUDG task, the inability to access data across different domains makes it even more challenging to obtain the global sample distribution necessary to address the long-tail problem. Specifically, when attempting to solve the long-tail issue from the perspective of a single client, it may encounter situations where the tail region of the client lies within the head sample area of the global distribution, thus failing to ensure consistency in cross-client model updates during aggregation. For example, as illustrated in Fig. \ref{fig:FedHUG}(c), the tail region of the BUAA precisely constitutes the head regions of the PURE and VIPL distributions. Furthermore, due to the inherent privacy of physiological data, methods \cite{du2024simpro} relying on pseudo-labels to address the problem of unsupervised data imbalance are also inapplicable to RPM FUDG.

To implicitly resolve this without transmitting privacy-sensitive information, we propose the Global Distribution-aware Learning Controller (GDLC). It models the abstract semantic distribution on a high-dimensional unit hypersphere and adaptively reweighs local training based on global distribution feedback. Inspired by \cite{du2024probabilistic}, we project the extracted feature vector of each sample onto the unit hypersphere as $v \in \mathbb{S}^{d-1}$. The distribution of these features is explicitly modeled using a von Mises–Fisher distribution \cite{mardia2009directional}, parameterized by a mean direction $\mu$ and a concentration parameter $\kappa$.

At the end of each local training round, client $i$ constructs its dataset-level statistics by aggregating the features of all its $N_i$ local samples. To ensure a shared representation space across clients, these features are extracted using a frozen copy of the same global encoder $f_{\theta^t}$ received at the beginning of round $t$, i.e., $v_i^n=f_{\theta^t}(x_i^n)/\|f_{\theta^t}(x_i^n)\|_2$. The maximum likelihood estimation for the local mean direction $\hat{\mu}_i$ and the mean resultant length $\bar{R}_i$ are derived as:

\begin{equation}
\hat{\mu}_i = \frac{\sum_{n=1}^{N_i} v_i^n}{\left\| \sum_{n=1}^{N_i} v_i^n \right\|_2}, \quad \bar{R}_i = \frac{\left\| \sum_{n=1}^{N_i} v_i^n \right\|_2}{N_i}
\label{eq: MLE_mu}
\end{equation}

To avoid numerical inversion involving complex Bessel functions, we adopt Banerjee's efficient approximation \cite{mardia2009directional} to estimate the concentration parameter inline as $\hat{\kappa}_i \approx \bar{R}_i\,(d - \bar{R}_i^2)/(1 - \bar{R}_i^2)$. 

Crucially, instead of exposing local features or communicating at the batch level, clients only upload the dataset-level parametric statistics $\{\hat{\mu}_i, \bar{R}_i, N_i\}$ to the server per communication round. This introduces negligible communication overhead and perfectly circumvents feature inversion attacks. The server then estimates the current-round global distribution $\{\mu_{global}^t, \kappa_{global}^t\}$ by aggregating the incoming statistics from all $K$ clients:

\begin{equation}
\begin{aligned}
\mathbf{r}_{global}^t
&= \sum_{i=1}^{K} N_i\bar{R}_i\hat{\mu}_i, \\
\mu_{global}^t
&= \frac{\mathbf{r}_{global}^t}{\left\|\mathbf{r}_{global}^t\right\|_2}, \\
\bar{R}_{global}^t
&= \frac{\left\|\mathbf{r}_{global}^t\right\|_2}{\sum_{i=1}^{K}N_i}
\end{aligned}
\label{eq: dynamic_update}
\end{equation}

where $\mathbf{r}_{global}^t$ is the pooled resultant vector, and $\kappa_{global}^t$ is updated using $\bar{R}_{global}^t$.

The updated global parameters are broadcast back to the clients. During the subsequent local training phase, we shift our focus from the dataset level to the mini-batch level to enable dynamic penalty weighting. For each training mini-batch $\mathcal{B}$ on the client, we swiftly estimate the batch-level direction $\hat{\mu}_{\mathcal{B}}$ and concentration $\hat{\kappa}_{\mathcal{B}}$. To quantify how severely the current local batch deviates from the global distribution, the client computes the Kullback-Leibler (KL) divergence between its batch vMF distribution and the global one:

\begin{equation}
D_{KL}(\mathcal{B}) = \log C_d(\hat{\kappa}_{\mathcal{B}}) - \log C_d(\kappa_{global}^t) + \left( \hat{\kappa}_{\mathcal{B}} - \kappa_{global}^t (\mu_{global}^t)^\top \hat{\mu}_{\mathcal{B}} \right) \cdot A_d(\hat{\kappa}_{\mathcal{B}})
\end{equation}

A pronounced $D_{KL}(\mathcal{B})$ explicitly reveals that the current mini-batch is populated by minority physiological states located in the global tail. However, directly applying unbounded divergence scores for loss re-weighting is highly susceptible to gradient explosion and training instability. To safely calibrate the optimization landscape, we project the divergence into a bounded importance factor $V_{KL} \in (1,2)$ via a parameterized Sigmoid gating function:

\begin{equation}
V_{KL} = 1 + \frac{1}{1 + \exp\left(-\frac{D_{KL}(\mathcal{B})-\sigma}{\gamma}\right)}
\label{eq: V_kl}
\end{equation}

Here, $\sigma$ represents the horizontal shift parameter that adjusts the sensitivity interval of the weighting, and $\gamma$ controls the steepness. This $V_{KL}$ directly scales the unsupervised learning loss of the current batch $\mathcal{B}$, effectively preventing the global representation from collapsing into the head distribution and ensuring robust generalization across imbalanced, unannotated domains. Notably, scaling up learning weights for tail samples risks amplifying local domain noise. FedHUG resolves this via a dual-level constraint: while GDLC drives local tail-state exploration at the micro-level, MBA serves as a global safety net. If a client overfits to domain artifacts during exploration, its degraded spatio-temporal consistency inherently suppresses its server aggregation weight. This interplay ensures the representation space is governed by intrinsic physiological rhythms, safely highlighting true rare states.

\subsection{Parameter Update in FedHUG}
\label{subsection: Federated Heterogeneous Unsupervised Generalization}


The overall procedure of FedHUG is summarized in Algorithm \ref{alg:fedhug}. Initially, a global model $\theta^0$ is supervisedly pre-trained to establish physiological priors. In each communication round, the server broadcasts the current model and global distribution parameters to all clients. During local training on unlabeled data $\mathcal{D}_i$, clients optimize a composite unsupervised loss (comprising bandwidth, sparsity, and variation constraints \cite{speth2023non}). Crucially, the GDLC mechanism dynamically scales this loss at the mini-batch level using the penalty $V_{KL}$, explicitly enhancing tail-sample exploration. Post-training, each client evaluates its translation-invariant consistency score $s_i$ and dataset-level statistics $\{\hat{\mu}_i, \bar{R}_i, N_i\}$. Finally, these comprehensive metrics, alongside the updated local weights, are uploaded to the server to simultaneously execute the global distribution update and the MBA, formulating the generalized model for the subsequent round.

\begin{algorithm}[H]
\caption{FedHUG Framework}
\label{alg:fedhug}
\KwIn{Rounds $T$, clients $\{\mathcal{D}_i\}_{i=1}^K$, learning rate $\eta$}
\KwOut{Global model $\theta^T$}

Server initializes global model $\theta^0$, and distribution parameters $\mu_{global}^0, \kappa_{global}^0$\;

\For{$t = 0$ \KwTo $T-1$}{
    \ForEach{client $i \in \{1, \dots, K\}$ \textbf{in parallel}}{
        Sync $\theta_i^t \leftarrow \theta^t$\;
        \ForEach{mini-batch $\mathcal{B} \subset \mathcal{D}_i$}{
            Estimate $\hat{\mu}_{\mathcal{B}}, \hat{\kappa}_{\mathcal{B}}$, and calculate $V_{KL}$ via Eq. (\ref{eq: V_kl}) \textit{(set $V_{KL}=1$ if $t=0$)}\;
            Local update: $\theta_i^t \leftarrow \theta_i^t - \eta \cdot V_{KL} \nabla \mathcal{L}(\theta_i^t; \mathcal{B})$\;
        }
        Compute SSM score $s_i$ (Eq. \ref{eq:COSINE}) and dataset-level stats $\{\hat{\mu}_i, \bar{R}_i\}$ (Eq. \ref{eq: MLE_mu})\;
        Upload $\{\theta_i^t, s_i, \hat{\mu}_i, \bar{R}_i, N_i\}$ to Server\;
    }
    Server updates global distribution $\{\mu_{global}^{t+1}, \kappa_{global}^{t+1}\}$ via Eq. (\ref{eq: dynamic_update})\;
    Server aggregates global model $\theta^{t+1}$ via MBA Eq. (\ref{eq:mba})\;
}
\Return $\theta^T$
\end{algorithm}

\section{Experiments}
\subsection{Experimental Setup}
\textbf{Datasets and Evaluation Protocol.}
We carefully select five rPPG datasets (\textbf{SCAMPS} \cite{mcduff2022scamps}, \textbf{PURE} \cite{stricker2014non}, \textbf{BUAA} \cite{xi2020image}, \textbf{MMPD} \cite{tang2023mmpd}, \textbf{VIPL} \cite{niu2019vipl}) and two mmWave datasets (\textbf{EquiPlet} \cite{vilesov2022blending}, and \textbf{PhysDrive} \cite{wang2026physdrive}), which include various conditions, to benchmark the large-scale FUDG protocol.

In implementing the FUDG protocol, we validate the RGB and mmWave scenes separately. For pre-training, we select a dataset  \(D_{pretrain}\) with abundant, evenly distributed samples and deploy the initial model on the server. The remaining dataset \(D_{source}\) is placed on independent clients with no data sharing. Then we separate the source domains' data into the train and validation set (4:1). After multi-round communications on the train set, we tune the model's hyper-parameters based on performance on the validation set, and lastly test the model on a target dataset \(D_{target}\) as an unseen domain. We test five times with different random seeds. We use MAE and its standard deviation (SD), RMSE, and Pearson Correlation Coefficient (p), then conduct a paired t-test between the best and second-best models.

\noindent\textbf{Implementation Details.}
Our proposed method is implemented under the Pytorch framework. Following Section \ref{subsection: Multi-Modal Representation and Physiological Augmentation}, facial videos are converted into STMaps $x_i^n \in \mathbb{R}^{256 \times 25 \times 3}$ and augmented spatio-temporally to yield $a_i^n$. Similarly, mmWave data is processed into range matrices $\in \mathbb{R}^{256 \times 5 \times 2}$ within a 1.25m region of interest \cite{wu2025cardiacmamba}. To strictly preserve privacy, the rPPG backbone (PhysMLE \cite{wang2025physmle}) is pre-trained on synthetic SCAMPS, while the mmWave backbone \cite{vilesov2022blending} is pre-trained on EquiPlet. Federated optimization runs for 100 communication rounds. We utilize the Adam optimizer (lr=$10^{-4}$, batch size=$100$) and empirically set $\tau=0.1, \beta=0.07, \gamma=0.5, \sigma=50$.

For the rPPG experiments, SiNC \cite{speth2023non} is used as the default local unsupervised objective. The same objective is used for all FL baselines to ensure a consistent comparison of federated optimization strategies.

\subsection{Quantitative Results}
\noindent\textbf{HR Estimation in rPPG.} Table~\ref{tabel: HR Estimation} reports the results on the four rPPG target datasets. The traditional methods, CHROM \cite{de2013robust} and POS \cite{wang2016algorithmic}, are directly evaluated on each target domain. The supervised DL baselines, PhysNet \cite{yu2019remote}, PhysFormer++ \cite{yu2023physformer}, RhythmFormer \cite{zou2025rhythmformer}, and PhysMLE \cite{wang2025physmle}, are pre-trained on SCAMPS and then evaluated on the target domains without further training. For the FL comparisons, FedHUG and the FL baselines \cite{mcmahan2017communication, li2020federated, hu2024fedmut, liao2024rethinking, pourpanah2024federated} share the PhysMLE backbone and use SiNC as the local unsupervised objective. These methods are evaluated following the FUDG protocol.

In line with previous works \cite{wang2023hierarchical}, DL methods perform better than traditional methods in most cases. Notably, the generalization gap between standard DL and FL models underscores the necessity of unsupervised adaptation following pre-training to mitigate domain shift. FedHUG significantly outperforms existing baselines through effective unsupervised joint learning. Particularly, we observed that, in PURE and BUAA, there was the largest performance improvement (MAE decreased by 9.6\% and 11.6\%) compared to the best existing FL method (i.e., FedGaLA). These more stable datasets benefit most from FedHUG, reflecting its ability to reduce bias and capture more semantic information.

\begin{table}[htbp]
\setlength{\tabcolsep}{1pt}
\renewcommand{\arraystretch}{1.2}
\centering
\scriptsize
\caption{HR estimation results under FUDG protocol in the rPPG task. ‘Base’ and ‘\(^+\)’ refer to the pre-trained model that is applied directly to the target without unsupervised training. ‘\(^*\)’ means it uses the same unsupervised loss and backbone as FedHUG.  The \textbf{bold} texts show the significantly (p-value $<$ 0.05) best result within each column with the paired-t test.}
\resizebox{\textwidth}{!}{
\begin{tabular}{lcccccccccccccccccccc}
\toprule
  & \multicolumn{4}{c}{\textbf{VIPL}} 
  & \multicolumn{4}{c}{\textbf{PURE}} 
  & \multicolumn{4}{c}{\textbf{BUAA}} 
  & \multicolumn{4}{c}{\textbf{MMPD}} \\
  \cmidrule(lr){2-5} \cmidrule(lr){6-9} \cmidrule(lr){10-13} \cmidrule(lr){14-17}
\textbf{Method} 
  & \textbf{MAE↓} & \textbf{SD↓} & \textbf{RMSE↓} & \textbf{p↑}
  & \textbf{MAE↓} & \textbf{SD↓} & \textbf{RMSE↓} & \textbf{p↑}
  & \textbf{MAE↓} & \textbf{SD↓} & \textbf{RMSE↓} & \textbf{p↑}
  & \textbf{MAE↓} & \textbf{SD↓} & \textbf{RMSE↓} & \textbf{p↑} \\
\midrule

CHROM \cite{de2013robust} 
  & 11.44 & 15.19 & 16.97 & 0.28 
  & 9.79  & 11.96 & 12.76 & 0.37 
  & 6.09  & 8.03  & 8.29  & 0.51 
  & 13.63 & 16.23 & 18.75 & 0.08 \\

POS \cite{wang2016algorithmic} 
  & 14.59 & 18.11 & 21.26 & 0.19 
  & 9.82  & 12.85 & 13.44 & 0.34 
  & 5.04  & 6.54  & 7.12  & 0.63 
  & 13.84 & 17.07 & 19.18 & 0.15 \\  

\midrule 

PhysNet\(^+\) \cite{yu2019remote} 
  & 13.53 & 15.13 & 16.46 & 0.35 
  & 18.12 & 22.71 & 24.32 & 0.43 
  & 7.21  & 7.40  & 8.12  & 0.58 
  & 19.11 & 24.17 & 26.29 & 0.04 \\

Physformer++\(^+\) \cite{yu2023physformer} 
  & 12.07 & 14.34 & 15.38 & 0.32 
  & 15.48 & 19.50 & 24.63 & 0.46 
  & 6.52  & 8.47  & 8.51  & 0.55 
  & 16.37 & 19.65 & 21.82 & 0.07 \\

RhythmFormer\(^+\) \cite{zou2025rhythmformer} 
  & 10.22 & 12.04 & 12.98 & 0.39 
  & 6.47  & 8.02  & 9.45  & 0.61 
  & 4.78  & 6.03  & 6.16  & 0.64 
  & 13.24 & 15.98 & 16.53 & 0.10 \\

PhysMLE\(^+\) \cite{wang2025physmle} 
  & 9.72  & 10.57 & 12.89 & 0.37 
  & 7.31  & 8.16  & 9.47  & 0.63 
  & 5.23  & 6.13  & 6.32  & 0.72 
  & 12.98 & 13.86 & 15.49 & 0.09 \\

\midrule

FedAvg\(^*\) \cite{mcmahan2017communication}  
  & 6.97  & 9.04  & 9.35  & 0.49 
  & 6.81  & 7.96  & 9.11  & 0.66 
  & 4.02  & 4.99  & 5.30  & 0.80 
  & 9.93  & 11.83 & 11.97 & 0.12 \\

FedProx\(^*\) \cite{li2020federated} 
  & 6.85  & 8.96  & 9.23  & 0.52 
  & 6.59  & 7.91  & 8.77  & 0.64 
  & 4.13  & 4.82  & 5.24  & 0.85 
  & 9.89  & 11.85 & 11.89 & 0.11 \\

FedMut\(^*\) \cite{hu2024fedmut} 
  & 6.92  & 9.01  & 9.16  & 0.50 
  & 6.71  & 7.85  & 8.57  & 0.64 
  & 4.37  & 4.84  & 5.63  & 0.84 
  & 9.94  & 11.74 & 12.01 & 0.09 \\

FedU2\(^*\) \cite{liao2024rethinking} 
  & 6.89  & 8.93  & 9.27  & 0.51 
  & 6.44  & 7.88  & 8.83  & 0.62 
  & 4.21  & 4.87  & 5.04  & 0.82 
  & 9.91  & 11.71 & 12.29 & 0.10 \\

FedGaLA\(^*\) \cite{pourpanah2024federated} 
  & 6.84  & 8.96  & 9.19  & 0.52 
  & 6.37  & 7.83  & 8.42  & 0.66 
  & 3.98  & 4.68  & 4.94  & 0.81 
  & 9.93  & 11.67 & 12.13 & 0.14 \\

\midrule 

FedHUG w/o MBA 
  & 6.89  & 9.01  & 9.26  & 0.53 
  & 6.19  & 7.86  & 8.24  & 0.69 
  & 3.68  & 4.59  & 4.88  & 0.83 
  & 9.78  & 11.75 & 11.93 & 0.15 \\

FedHUG w/o GDLC 
  & 6.79  & 8.98  & 9.09  & 0.50 
  & 6.18  & 7.85  & 8.21  & 0.68 
  & 3.58  & 4.53  & 4.93  & 0.81 
  & 9.75  & 11.69 & 11.88 & 0.14 \\

\textbf{FedHUG}  
  & \textbf{6.73}\(^*\) & \textbf{8.90}\(^*\) & \textbf{9.06}\(^*\) & \textbf{0.55}\(^*\) 
  & \textbf{5.76}\(^*\) & \textbf{7.80}\(^*\) & \textbf{8.08}\(^*\) & \textbf{0.69}\(^*\) 
  & \textbf{3.52}\(^*\) & \textbf{4.08}\(^*\) & \textbf{4.77}\(^*\) & \textbf{0.85}\(^*\) 
  & \textbf{9.73}\(^*\) & \textbf{11.64}\(^*\) & \textbf{11.87}\(^*\) & \textbf{0.16}\(^*\) \\

\bottomrule
\end{tabular}}
\label{tabel: HR Estimation}
\end{table}

\noindent\textbf{Testing on Source Domains.} While FUDG prioritizes unknown domain generalization, maintaining source domain performance is vital to prevent catastrophic forgetting during deployment. Thus, we cancel the target domain evaluation but test on each source domain after FUDG training. As shown in Tab. \ref{tabel: cross-dataset}, FedHUG consistently outperforms all baselines, demonstrating robust generalization without compromising local client accuracy.

\begin{table}[htbp]
\setlength{\tabcolsep}{3pt} 
\renewcommand{\arraystretch}{1.2}
\centering
\scriptsize
\caption{Results on source domains. The model is pre-trained on SCAMPS and federated unsupervised trained on four datasets.}
\resizebox{\textwidth}{!}{
\begin{tabular}{lcccccccccccccccc}
\toprule
  &\multicolumn{4}{c}{\textbf{VIPL}} 
  &\multicolumn{4}{c}{\textbf{PURE}} 
  &\multicolumn{4}{c}{\textbf{BUAA}} 
  &\multicolumn{4}{c}{\textbf{MMPD}} \\
  \cmidrule(lr){2-5} \cmidrule(lr){6-9} \cmidrule(lr){10-13} \cmidrule(lr){14-17}
\textbf{Method} 
  & \textbf{MAE↓}  & \textbf{SD↓} & \textbf{RMSE↓} & \textbf{p↑}    
  & \textbf{MAE↓}  & \textbf{SD↓} & \textbf{RMSE↓} & \textbf{p↑}    
  & \textbf{MAE↓}  & \textbf{SD↓} & \textbf{RMSE↓} & \textbf{p↑}    
  & \textbf{MAE↓}  & \textbf{SD↓} & \textbf{RMSE↓}& \textbf{p↑ } \\
\midrule

FedAvg \cite{mcmahan2017communication}  
  & 6.86 & 8.98 & 9.39 & 0.52 
  & 5.61 & 7.71 & 8.09 & 0.57 
  & 3.80 & 4.65 & 4.89 & 0.82 
  & 9.79 & 11.24 & 11.79 & 0.13 \\

FedProx\(^*\) \cite{li2020federated} 
  & 6.78 & 8.73 & 9.26 & 0.54 
  & 5.51 & 7.66 & 7.94 & 0.59 
  & 3.35 & 4.81 & 4.81 & 0.84 
  & 9.69 & 11.07 & 11.60 & 0.14 \\

FedMut\(^*\) \cite{hu2024fedmut} 
  & 6.82 & 8.52 & 9.41 & 0.48 
  & 5.88 & 8.25 & 8.96 & 0.58 
  & 3.68 & 4.74 & 5.05 & 0.83 
  & 9.77 & 11.28 & 11.93 & 0.11 \\

FedU2 \cite{liao2024rethinking}  
  & 6.89 & 8.48 & 9.20 & 0.53 
  & 5.78 & 7.41 & 8.15 & 0.59 
  & 3.49 & 4.61 & 4.72 & 0.82 
  & 9.85 & 10.88 & 11.59 & 0.14 \\

FedGaLA \cite{pourpanah2024federated} 
  & 6.61 & 8.38 & 9.13 & 0.55 
  & 5.61 & 7.29 & 7.74 & 0.56 
  & 3.29 & 4.60 & 4.77 & 0.85 
  & 9.70 & 10.69 & 11.55 & 0.16 \\

\midrule 

\textbf{FedHUG}  
  & \textbf{6.55}\(^*\) & \textbf{8.17}\(^*\) & \textbf{8.98}\(^*\) & \textbf{0.57}\(^*\) 
  & \textbf{5.29}\(^*\) & \textbf{7.03}\(^*\) & \textbf{7.47}\(^*\) & \textbf{0.60} \(^*\)
  & \textbf{3.19} \(^*\)& \textbf{4.38} \(^*\)& \textbf{4.58} \(^*\)& \textbf{0.87} \(^*\)
  & \textbf{9.64} \(^*\)& \textbf{10.36} \(^*\)& \textbf{11.48} \(^*\)& \textbf{0.17}\(^*\) \\

\bottomrule
\end{tabular}}
\label{tabel: cross-dataset}
\end{table}

\noindent\textbf{Cross-Modality Verification in mmWave.} To verify FedHUG's modality-agnostic generalization, we extend the FUDG protocol to mmWave signals. Using the PhysDrive \cite{wang2026physdrive}, we partition data by vehicular road conditions (Flat \& Unobstructed, Flat \& Congested, Bumpy \& Congested), designating one condition as the unseen target domain and the others as source domains. As shown in Tab. \ref{tabel: mmwave}, FedHUG consistently outperforms existing FL baselines across these dynamic environments. This confirms that our framework operates purely on abstract semantic spaces, seamlessly generalizing across diverse sensing modalities.

\begin{table}[htbp]
\setlength{\tabcolsep}{3pt}
\renewcommand{\arraystretch}{1.2}
\centering
\scriptsize
\caption{HR estimation results with mmWave under FUDG.}
\setlength{\tabcolsep}{2.75pt}
\begin{tabular*}{\textwidth}{@{\extracolsep{\fill}}lcccccccccccc}
\toprule
  & \multicolumn{4}{c}{\textbf{Flat \& Unobstructed}} 
  & \multicolumn{4}{c}{\textbf{Flat \& Congested}} 
  & \multicolumn{4}{c}{\textbf{Bumpy \& Congested}}\\
  \cmidrule(lr){2-5} \cmidrule(lr){6-9} \cmidrule(lr){10-13}
\textbf{Method} 
  & \textbf{MAE↓} & \textbf{SD↓} & \textbf{RMSE↓} & \textbf{p↑}    
  & \textbf{MAE↓} & \textbf{SD↓} & \textbf{RMSE↓} & \textbf{p↑}    
  & \textbf{MAE↓} & \textbf{SD↓} & \textbf{RMSE↓} & \textbf{p↑}  \\
\midrule

Base 
  & 11.21 & 14.66 & 15.91& 0.09 
  & 11.45 &14.35 &15.27 &0.12 
  & 11.56 & 14.98 &15.73 &0.10 \\

FedAvg 
  & 10.03 & 12.74 &13.07 & 0.19 
  & 9.89 &12.60 & 12.56 & 0.15 
  & 10.21 &12.78 & 12.84 & 0.16 \\

FedProx 
  & 9.79 & 12.60 & 12.85 & 0.23 
  & 9.54 & 12.42 & 13.66 & 0.21 
  & 9.95 &12.83 & 12.90 & 0.19 \\

FedMut 
  & 9.90 & 11.87 & 12.67 & 0.22 
  & 10.19 & 11.97 & 13.12 & 0.18
  & 9.93 & 12.41 & 12.49 & 0.23 \\

FedU2 
  & 9.87 & 11.49 & 11.54 & 0.25 
  & 9.97 & 10.66 & 10.93 & 0.24 
  & 10.22 &11.06 & 12.41 & 0.23 \\

FedGaLA 
  & 9.63 & 10.36 & 10.95& 0.27 
  & 9.75 &10.64 & 10.83 & 0.24 
  & 9.64 & 10.27 & 10.63& 0.23 \\
\midrule 

\textbf{FedHUG}  
  & \textbf{9.38}\(^*\) & \textbf{9.96}\(^*\) & \textbf{10.51}\(^*\)& \textbf{0.31} \(^*\)
  & \textbf{9.31} \(^*\)&\textbf{9.40} \(^*\)& \textbf{10.42}\(^*\)& \textbf{0.28} \(^*\)
  & \textbf{9.45} \(^*\)&\textbf{9.83} \(^*\)& \textbf{10.28}\(^*\)& \textbf{0.27} \(^*\)\\
\bottomrule
\end{tabular*}
\label{tabel: mmwave}
\end{table}

\noindent\textbf{Ablation Study.} Table~\ref{tabel: HR Estimation} evaluates the individual and combined contributions of MBA and GDLC. All averages reported here assign equal weight to the four rPPG target datasets. Compared with FedAvg, adding MBA alone (FedHUG w/o GDLC) reduces average MAE by 5.16\%, while adding GDLC alone (FedHUG w/o MBA) reduces it by 4.29\%. Combining both components further reduces average MAE by 2.13\% relative to the MBA-only variant and by 3.01\% relative to the GDLC-only variant. Adding GDLC to the MBA-only variant also increases the mean Pearson correlation coefficient from 0.5325 to 0.5625. These results show that each component improves average performance and that their combination achieves the lowest average MAE among the evaluated variants. Separately, compared with the SCAMPS-pre-trained PhysMLE model evaluated directly on the target domains without federated adaptation, FedHUG reduces average MAE by 26.96\%. This comparison measures the benefit of the complete federated adaptation procedure over direct inference from the pre-trained model.

\begin{figure}[ht] 
    \centering
    \includegraphics[width=0.90\textwidth]{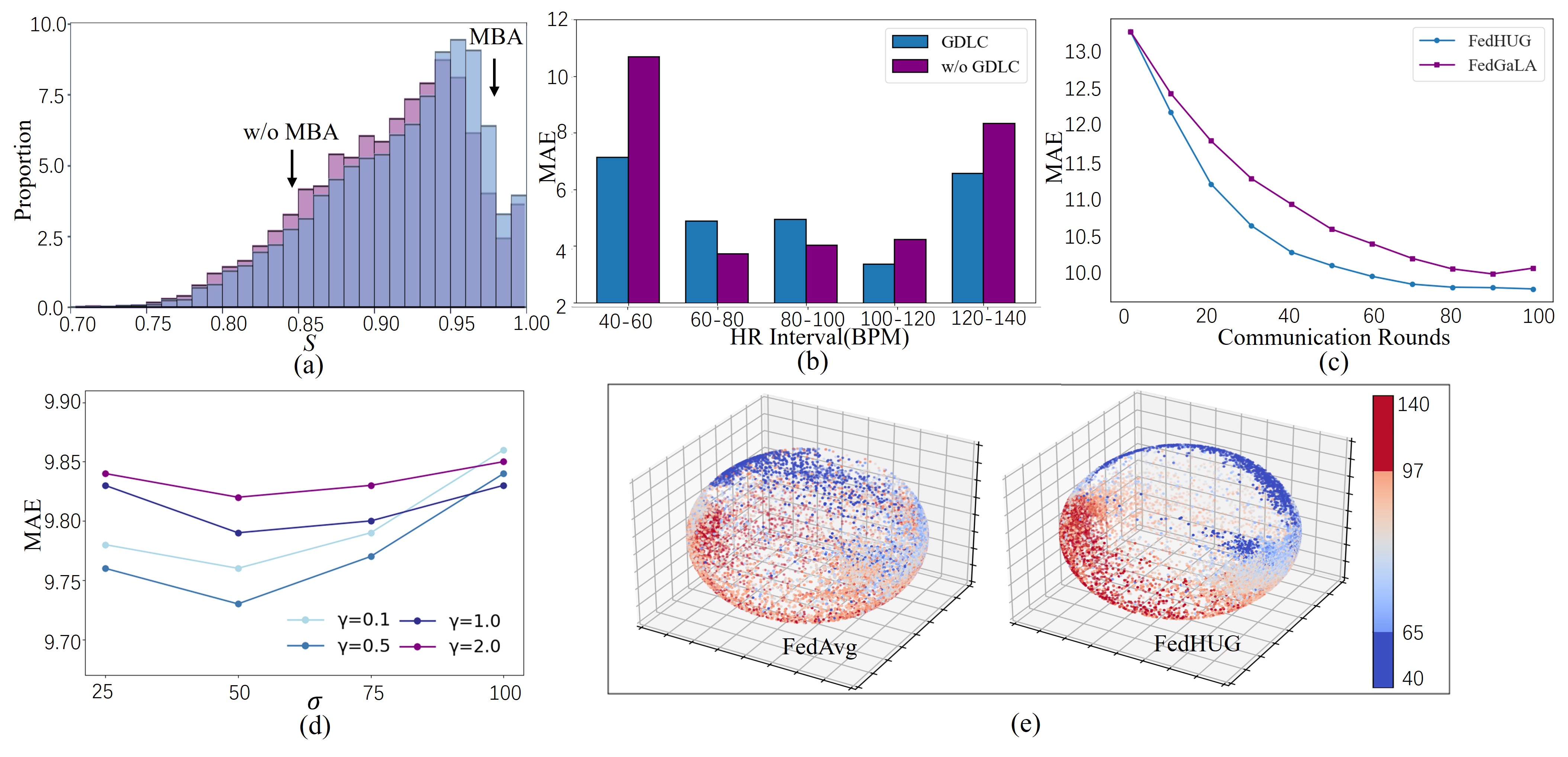} 
    \caption{ Illustration of the distribution of \(s\) in PURE before and after applying the MBA method (a), and MAE at different HR intervals before and after the use of GDLC (b). MAE on MMPD across communication rounds under the FUDG protocol (c) . The MAE variation across different communication rounds of the MMPD dataset under the FUDG protocol, the impact of the \(\gamma\) and \(\sigma\) on the model’s performance (d). BUAA's feature distribution on unit sphere (e).}
    \label{fig: visualizing of Ablation}
\end{figure}

\subsection{Visualization Results}

\noindent\textbf{Bias Elimination.} To verify whether the model effectively moves towards semantic invariance points, we visualize the distribution of \(s\) before and after applying MBA. In Fig. \ref{fig: visualizing of Ablation} (a), we observe that without the MBA, the \(s\) distribution has a larger proportion at low-similarity values, indicating that the model exhibits poor generalization ability on target datasets and is significantly impacted by bias. By contrast, FedHUG shows a notable improvement, and most instances present a high \(s\) in the distribution, demonstrating that it effectively reduces bias from the source domains and emphasizes the need for tailored methods in RPM.

\noindent\textbf{Phys-label Balance Analysis.} We compared MAE across different HR segments (Fig. \ref{fig: visualizing of Ablation}(b)). On the PURE, FedHUG with GDLC showed significant improvement on tail intervals (\(<\)60, \(\geq\)100), and performed comparably on head data, demonstrating its efficacy in alleviating long-tail distribution bias.

\noindent\textbf{Communication Round.} To investigate the model performance of FedHUG compared to the best existing method, FedGaLA, we recorded the changes in MAE values of the target dataset under the FUDG framework across different communication rounds in Fig. \ref{fig: visualizing of Ablation}(c). In this experiment, a local model was trained in each communication round. Taking MMPD as an example, we observed that FedHUG exhibited a faster MAE decline, with better model performance.

\noindent\textbf{Impacts of the Hyperparameter.} We tested our model's performance on the MMPD domain for different \(\sigma\) and \(\gamma\). As shown in Fig. \ref{fig: visualizing of Ablation}(d), the model performs best with (\(\sigma = 50\), \(\gamma = 0.5\)). Performance declines when \(\sigma\) is too high or low, as GDLC struggles to detect anomalous training batches. Additionally, \(\gamma\) affects the effectiveness of GDLC once activated.

\noindent\textbf{Visualization of Feature Distribution.} To visualize how FedHUG handles long-tail distributions, we project semantic features onto a 3D unit sphere via PCA (Fig. \ref{fig: visualizing of Ablation}(e)). Points are color-coded by HR, with extreme tails (<10th and >90th percentiles) highlighted in darker shades. While standard FedAvg features are chaotically entangled, FedHUG yields a highly structured topology. Tail samples are distinctly clustered and pushed in opposite directions for high and low HRs. This perfectly visualizes our premise in Sec. \ref{subsection: Global Distribution Aware Learning Controller}: GDLC actively disentangles the representation space, successfully sensitizing the network to rare physiological semantics.

\section{Conclusion}

This work investigated federated unsupervised domain generalization (FUDG) for remote physiological measurement and proposed FedHUG to address heterogeneous feature and physiological label distributions. By combining Minimal Bias Aggregation with a Global Distribution-aware Learning Controller, FedHUG improves learning from unlabeled client data while keeping raw measurements local. Experiments on RGB video and mmWave radar datasets demonstrate improved generalization to unseen domains, while the ablation and visualization results support the complementary roles of the two components. These findings provide a basis for further research on physiological representation learning under decentralized, unlabeled, and heterogeneous data conditions.

Nevertheless, a gap remains between cross-dataset evaluation and long-term real-world deployment. The current experiments assess generalization across existing datasets and collection conditions, but do not establish whether these improvements can be sustained as user behavior, sensing conditions, and physiological states change over time. Longitudinal evaluation is therefore needed to assess the stability and practical applicability of FedHUG beyond the current benchmarks.

Future work will investigate individual-level FUDG, extending the evaluation from dataset-level domain differences to variation among individual users. This direction will examine whether a shared model can generalize to previously unseen individuals while accounting for differences in their physiological characteristics and recording conditions. We also plan to evaluate FedHUG in long-term deployment settings to better understand its behavior under evolving data distributions and identify the conditions under which federated unsupervised updates remain beneficial.


\section*{Declaration of generative AI and AI-assisted
technologies in the manuscript preparation process}

During the preparation of this manuscript, the authors used
OpenAI Codex to assist with language editing, organization,
and revision of technical explanations. The authors reviewed
and revised the AI-assisted content and take full responsibility
for the final manuscript.

\bibliographystyle{elsarticle-num}
\bibliography{main}
\end{document}